\documentclass[letterpaper, 10 pt, conference]{ieeeconf}  
\usepackage{graphicx}
\graphicspath{{./images/}}
\usepackage{algorithm, algorithmic, url}
\usepackage{booktabs}

\if CLASSOPTIONcompsoc
    \usepackage[caption=false, font=normalsize, labelfont=sf, textfont=sf]{subfig}
\else
    \usepackage[caption=false, font=footnotesize]{subfig}
\fi

\IEEEoverridecommandlockouts                              

\title{\LARGE \bf
Real-Time Object Detection and Recognition on Low-Compute Humanoid Robots using Deep Learning
}

\author{Sayantan Chatterjee, Faheem H. Zunjani, Souvik Sen and Gora C. Nandi
\thanks{* All the authors are with the Robotics and Artificial Intelligence Laboratory at the Indian Institute of Information Technology, Allahabad, Prayagraj (Uttar Pradesh), India.}
}

\begin{document}

\maketitle
\thispagestyle{empty}
\pagestyle{empty}

\begin{abstract}

We envision that in the near future, humanoid robots would share home space and assist us in our daily and routine activities through object manipulations. One of the fundamental technologies that need to be developed for robots is to enable them to detect objects and recognize them for effective manipulations and take real-time decisions involving those objects. In this paper, we describe a novel architecture that enables multiple low-compute NAO robots to perform real-time detection, recognition and localization of objects in its camera view and take programmable actions based on the detected objects. The proposed algorithm for object detection and localization is an empirical modification of YOLOv3, based on indoor experiments in multiple scenarios, with a smaller weight size and lesser computational requirements. Quantization of the weights and re-adjusting filter sizes and layer arrangements for convolutions improved the inference time for low-resolution images from the robot’s camera feed. YOLOv3 was chosen after a comparative study of bounding box algorithms was performed with an objective to choose one that strikes the perfect balance among information retention, low inference time and high accuracy for real-time object detection and localization. The architecture also comprises of an effective end-to-end pipeline to feed the real-time frames from the camera feed to the neural net and use its results for guiding the robot with customizable actions corresponding to the detected class labels.

\end{abstract}


\section{Introduction}
\label{introduction}
Humanoid robots are being extensively experimented and tested upon uncontrolled environments to enable them to perform conditional actions specific to visual cues. In order to invoke an immediate response to objects or gestures introduced in the frame, the robot needs to be equipped with the capability to rapidly detect and recognise the object within its field of view. Achieving an accurate and fast inference from an object detection algorithm remains an open challenge.

SoftBank Robotics' NAO\footnote{SoftBank Robotics NAO: \url{https://www.softbankrobotics.com/emea/en/nao}} robot is increasingly been used in research exploring human-humanoid interaction and its applications like interaction therapy \cite{nao_autism}, assisted-living \cite{nao_elderly} \cite{nao_elderly2} and investigating non-verbal cues \cite{nao_nonverbal_cues}. To facilitate integration of a humanoid like NAO in a real-world environment outside of a lab setting, it is important to increase its ability to interact with everyday objects in its surroundings.

The goal of this project is to enable NAO robot to detect, classify and localize objects in its field of view in real-time and take actions based on the detected objects. 
NAO, a programmable bipedal, humanoid robot, has had many iterations over the years featuring 14, 21 or 25 degrees of freedom through its movable joints, fully equipped with an inertial measurement unit, accelerometer, gyrometer, cameras and peripheral microphones that enable it to possess a general awareness of its environment.
In terms of a software stack, NAO operates on its own specialised distro of Linux termed \textit{NAOqi} accompanied by a software suite including a graphical programming tool called Choregraphe and an SDK. 


The starting point to work on this object detection problem, specific to the NAO robot's use case in consideration, is to analyse the comparative performance results of the state-of-the art algorithms at hand. The metrics for judging an algorithm are not quite as straightforward as it may seem --- several factors come into play other than the accuracy viz. training time, inference time, performance in crowded and underexposed images, etc. The three metrics that have been primarily focused on here are:

\begin{itemize}
    \item \textbf{Accurate object detection}: Recognizing the presence of a particular object as well as its identification
    \item \textbf{Accurate localization}: Accuracy of the rectangular bounding box around the detected object
    \item \textbf{Inference time}: The time interval elapsed between the capturing of a particular frame from the video stream and yielding the processed results
\end{itemize}

At present, there are multiple approaches to solve the problem of accuracy in object detection, as in \cite{detection}. Region-based neural networks like RCNN, Faster RCNN and RFCN can achieve pixel or sub-pixel perfect results due to multiple training layers, but the inference time for such networks is not small enough to evoke a real-time action by processing the video input from the robot's camera. Moreover, due to memory and storage limitations as well as constraints of processing power, most of these networks either fail to run, or severely throttle on the robot hardware, which lacks the cooling mechanism to smoothly function under such extensive load. Hence, a more light-weight algorithm is needed that is relatively less demanding computationally, yet yields fairly accurate and faster results.

Microsoft's COCO dataset \cite{coco} has become an industry standard for training object detection and localisation algorithms, and has been used as the data set to benchmark the aforementioned algorithms. 

Object detection is a field highly dominated by CNN-based models. Therefore, we tested variants of two state-of-the-art algorithms:

\begin{itemize}
    \item Region-Based CNN \cite{fast-rcnn} \cite{faster-rcnn} \cite{mask-rcnn}, including a fully convolutional variant \cite{rfcn}
    \item YOLO (Darknet) \cite{yolo} \cite{yolov3} \cite{yolo9000}
\end{itemize}

One of the main areas of focus was to measure the trade-offs required to gain real-time inference capabilities with minimal sacrifice to accuracy. Since, our goal is to perform real-time inference by utilizing the input from NAO robot’s camera (low compute robot), the performance of the model both in terms of latency as well as accuracy was measured. Based on the performance comparisons, we chose YOLO and modified it to work better with low-res images (480p stream from NAO robot’s camera). The resultant model has been described here. A complete pipeline was developed which takes input from NAO's camera, feeds it to the detection algorithm, takes the output of the algorithm and then decides on the actions to be taken by NAO based on the predicted labels. The pipeline does all this in real-time.

The rest of the paper is organized as follows. In section \ref{related_work}, the state-of-the-art and previous related work has been discussed. In section \ref{problem_definition}, the problem has been formalized and elaborated upon and the architecture proposed has been discussed in section \ref{methodology}. Section \ref{results} describes the results and compares the performance among the different algorithms. The future work has been discussed and proposed in the last section. 
\section{Related Work}
\label{related_work}

In our previous work \cite{intent_object_grasping}, we tackled the problem of determining the ideal grasp location for an object in the field of view of a robot, based on the intention for grasping that object. From that project, we realized the challenge of accurately detecting object boundaries for a robot when the object is present in a real-world environment with many other distracting objects in the field of view of a robot. The problem considered in this paper is related to the \textit{bounding box} problem in the field of object detection, whereby a rectangular region is drawn around the target object to mark its closest wrapped boundaries inside the frame of the image. A predictive \textit{label} is associated with the recognition of the object and is mentioned for each of such bounding boxes along with a confidence metric for said predictions. The prior state-of-the-art algorithms that deal with this problem have been broadly classified as follows:

\subsection{Region-Based Convolutional Neural Networks: two-stage approaches}

Region-Based CNNs are an extension of the work done in \cite{imagenet} to generalize object classification to object detection. R-CNN uses a region proposal method called \textit{selective search} to propose sections of the image which are then run through a conventional CNN to see whether the object actually is contained within the region. The accuracy of the box is improved by a linear regression to generate increasingly tighter bounding boxes. Improvements of R-CNNs that have been developed till now are Fast R-CNN, Faster R-CNN and Mask R-CNN.

Fast R-CNN is aimed at reducing the computation required by running the forward pass on the image only once per image instead of once per proposal and then reusing the computation for each proposed region. Faster R-CNN takes it further by getting rid of the selective search process and instead utilises the results of the forward pass of the CNN. Mask R-CNN builds on Faster R-CNN by creating a binary mask of 0 and 1s which corresponding to the prediction of whether an object is present within the mask or not. Thus Mask R-CNN is able to segmentation at a pixel level.

Region-based Fully Convolutional Network (R-FCN) is a faster solution that attempts to solve the dilemma between translation-invariance in image classification and translation-variance in object detection by using position-sensitive score maps. Unlike the R-CNN variants, R-FCN removes the fully connected layers after the position-sensitive Region of Interest (ROI) pooling from the Region Proposal Network (RPN), followed by class-agnostic bounding box regression.

\subsection{YOLO and its variants: one-stage approaches}

YOLO is a new approach to object detection, originally developed by Joseph Redmon et al. [11]. Prior work on object detection repurposes classifiers to perform detection. Instead, this approach frames object detection as a regression problem to spatially separated bounding boxes and associated class probabilities. A single neural network predicts bounding boxes and class probabilities directly from full images in one evaluation. Since the whole detection pipeline is a single network, it can be optimized end-to-end directly on detection performance. The unified architecture is much faster compared to industry standards. The base YOLO model processes images in real-time at 45 frames per second.

Compared to state-of-the-art detection systems, YOLO makes more localization errors but is less likely to predict false positives on the background. Finally, YOLO learns very general representations of objects. It outperforms other detection methods, including DPM and R-CNN, when generalizing from natural images to other domains like artwork. Since YOLO has faster inference times, this algorithm is possibly a good suit for real-time object detection.
\section{Problem Definition}
\label{problem_definition}

Given a real-time video stream, objects in the stream need to be detected and localized so that an agent like a robot can take actions based on the label and location of the object.

\subsection{Input}

The input to the neural network is a video stream from the robot’s camera feed.

\subsection{Output}

The output of the neural network will be the class label of the detected object and its location in the reference frame of the image which can then be translated to the real word frame using a depth sensor. The inferences made by the model can also be visualized by drawing bounding boxes over the detected objects captured from the robot’s perspective.

After inference, the robot is able to select from a range of programmable actions depending on the class label of the object.

\subsection{Constraints}

The quality of the input feed given to the neural network is less than ideal due to the extremely limited compute capabilities of the NAO robot. The processor is unable to provide a steady 30 fps feed at 480p so the video looks choppy at times.\footnote{According to the specifications provided by Softbank Robotics, the top camera of NAO can record upto 1080p video at 30 fps while the bottom camera caps off at 640x480 at 30 fps, which is less than 480p. The wireless connection latency imposes a further limit on the transfer of captured frames, thereby rendering the higher resolutions largely unusable for real-time processing.}

\section{Methodology}
\label{methodology}

\subsection{Comparing state-of-the-art bounding box detection algorithms}

A comparative study of the aforementioned state-of-the-art object localization (bounding box detection) algorithms from \cite{faster-rcnn}, \cite{rfcn}, \cite{yolov3} on the following metrics has been performed, the results of which are illustrated in Fig.~\ref{comparison}:

\begin{itemize}
    \item Accuracy of detection and recognition
    \item Training time
    \item Mean inference time
    \item Percentage of false positives in crowded images
\end{itemize}

Since there exists a severe compute constraint, i.e., the computation power of the NAO robot, the input to the network is choppy and of relatively low resolution. However, inference time needed is ideally under 1 second to enable real-time detection in the stream. Accuracy needed is high, keeping in mind the low resolution of the input stream for best performance when tested on the robot. So, after comparing the above-mentioned algorithms, we found that a modifying the YOLO architecture suited our use case the best. Of the compared models, YOLO provided the best trade-off between accuracy and fast inference time. The Region based CNN models while a lot slower were about the same in accuracy due to the lower resolution of the test images from the NAO robot.

\subsection{Modifying YOLO for the use-case of the robot}

\begin{figure}[htpb]
    \centering
    \includegraphics[width=\columnwidth]{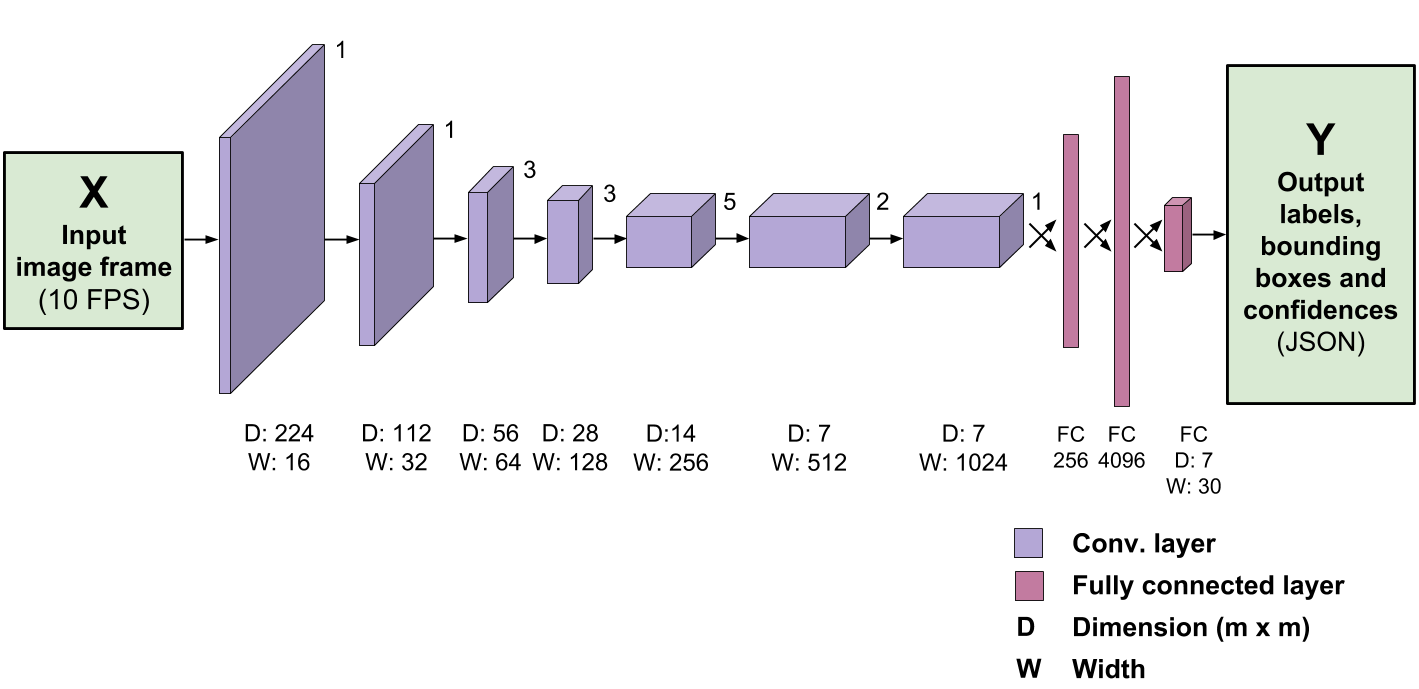}
    \caption{Modified network model for object recognition, segmentation and localization}
    \label{yolomodel}
\end{figure}

Based on the extensive guidelines and design principles stated in Cruz et al. \cite{guidelines}, the YOLO model has been modified to work faster on the lower resolution images being dealt with, as illustrated in Fig.~\ref{yolomodel}. Filter sizes of the second convolutional layer have been decreased to attain sub-second inference times with minimal loss of overall accuracy.

The two approaches that have been involved in modification are:

\subsubsection{Weight quantization}

In order to avoid a larger loading time and larger memory consumption, it is essential that the weight size of the neural network be as small as possible without a significant drop in overall accuracy. It is due to this reason that a quantization approach has been involved, in a similar vein to the earlier work in \cite{yolo-quantized} for YOLOv2, replacing floating-point computations by relatively faster integer computes. The benefits of converting 32-bit floats to 8-bit integers are two-fold: (a) a considerable reduction in bandwidth by reducing the total weight size of the network from 236.52MB to 94.89MB and faster inference times (mean gain of 2 ms), although partially sacrificing the accuracy (loss of 1.93\%).

\subsubsection{Reduction in the number of filters of the convolution layers}

In order to achieve the throughput required within the limited compute constraints of the NAO robot, the number of operations in forward passes need to be decreased. As per the design guidelines provided in \cite{guidelines}, several approaches viz. removing convolution layers, replacing the two fully connected layer with a single one, and reducing the width, i.e., the filter sizes of the convolution layers were attempted in the due course of a sequentially iterative Pareto optimization with number of layers, number of filters and filter width as the design criteria with the goal of minimizing the inference time while maximizing the accuracy.

Having analysed their performance comparisons, the best results next to YOLOv3-320 have been obtained by halving the width of the 1024 width layer and dropping the penultimate convolution layer from the original YOLOv3-320 model chosen, minimizing the accuracy loss in this step to 2.95\% while maximizing the gain in mean inference time to a further 3 ms.

In comparison to the optimised Darknet variant of the \textit{YOLOv3} model in \cite{yolo-optimised} 
having 18 convolutional layers, our network has 16 convolutional layers as illustrated in Fig.  \ref{yolomodel}, with each segment followed by a 2x2 \textit{maxpool} layer of stride 2. This is followed by 3 fully connected layers that yield the final output. The performance comparison of this modified network model relative to Faster R-CNN, R-FCN and YOLOv3 has been illustrated in Section \ref{results}.

\subsection{Connecting the NAO Robot to the pipeline}

The actual inference is too computationally demanding to happen on the NAO robot itself. An approach for designing a scalable neural network for bounding boxes has been elaborated upon in \cite{scalable}, referring to which our network has been separated from the NAO's CPU to have the computationally demanding convolution operations performed separately on a portable embedded system on which the trained model has been deployed, hereinafter referred to as the \textit{inference engine}, capable of meeting those hardware requirements so that the actual inferring can take place then and there in real-time. This inference engine may serve as a central hub to connect to multiple robots within the same household to provide a robust home assistance framework.

The desirables of the inference engine are twofold:
\begin{itemize}
    \item The inference engine should have minimal hardware, enough to support the steady computations required for running the deployed network without throttling, thereby reducing any significant addition of cost on the existing system.
    \item The engine should be connected to the NAO robot using the same wireless LAN connection to ensure low latency transfer to and from the engine.
\end{itemize}

The best cost-effective embedded system that satisfies the requirement of parallel processing with an embedded GPU is NVIDIA Jetson Nano\footnote{NVIDIA Jetson Nano: \url{https://developer.nvidia.com/embedded/jetson-nano-developer-kit}} that has 128 Cuda cores for efficient computing. Our deployed model performs efficiently on this machine without any noticeable throttling issues, and hence, this machine was a suitable choice for our inference engine on which the proposed optimized YOLOv3 model has been trained and deployed. It sends bounding box results with $label$, $confidence$, $topleft$ and $bottomright$ values on processing the input video frames.

\begin{figure}[htpb]
    \centering
    \includegraphics[width=\columnwidth]{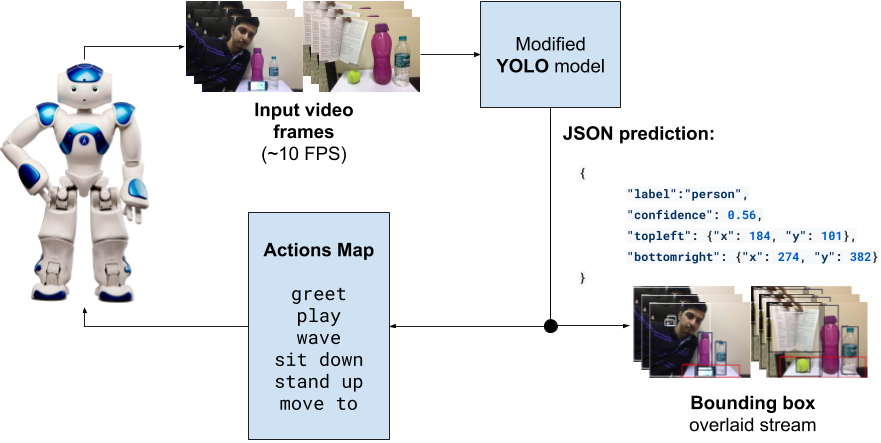}
    \caption{A complete overview of the system}
    \label{overview}
\end{figure}

Algorithm \ref{pipeline_algo} describes the algorithm developed for handling the stream and their inference in the pipeline. Each image captured from NAO's camera is sequentially fed to the neural network for inferring. The output of the neural network are the bounding box and the object class tag for each detected object in the image. This is later visualized by superimposing the bounding boxes on the input image to manually inspect the correctness of the prediction. These images can then be recombined to create a video stream of the inferring process and its outputs.

\begin{algorithm}[thpb]
    \caption{Algorithm for real-time object recognition and localization}
    \label{pipeline_algo}
    \begin{algorithmic}[1]
        \renewcommand{\algorithmicrequire}{\textbf{Input:}}
        \renewcommand{\algorithmicensure}{\textbf{Output:}}
        \REQUIRE Real-time head camera video feed from NAO at a frame rate of at least 10 fps
        \ENSURE Coordinates of detected bounding box diagonals, corresponding object labels and confidence values
        \STATE Initialise $processQueue$ to an empty queue
        \STATE Capture NAO robot's head camera video feed
        \STATE Compute frame rate $r$
        \STATE Branch the video feed to the inference engine and the monitoring stream
        \\ \textit{LOOP Process}
        \WHILE {frame capture is on}
            \FOR {each new frame $F$}
                \IF {inference engine is active}
                    \STATE Store frame $F$ in $processQueue$
                \ELSE
                    \STATE Run network model with input $F$ for inferring
                    \STATE Send output labels, confidence values and bounding box coordinates to NAO
                    \STATE Send $F$ overlaid with model's outputs to process monitoring stream
                \ENDIF
                \IF {($processQueue.length \ge max\{5,r\}$)}
                    \STATE Drop all frames in $processQueue$
                \ENDIF
            \ENDFOR
        \ENDWHILE
    \end{algorithmic}
\end{algorithm}

The pipeline which takes the real-time video stream is connected to the NAO robot and the video stream from NAO's cameras is fetched via sockets and fed to the neural network described in Fig.~\ref{yolomodel}. Based on the inference result, the NAO robot can also be given the command to execute a particular action or a sequence of action utilizing an extensible actions API.

\begin{figure}[htpb]
    \centering
  \subfloat[\label{exp-a}]{%
       \includegraphics[width=0.49\linewidth]{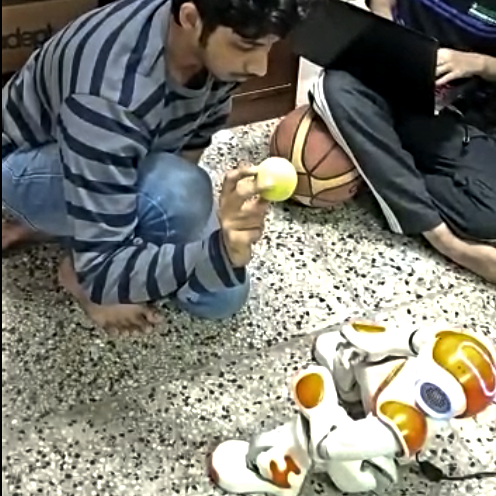}}
    \hfill
  \subfloat[\label{exp-b}]{%
        \includegraphics[width=0.49\linewidth]{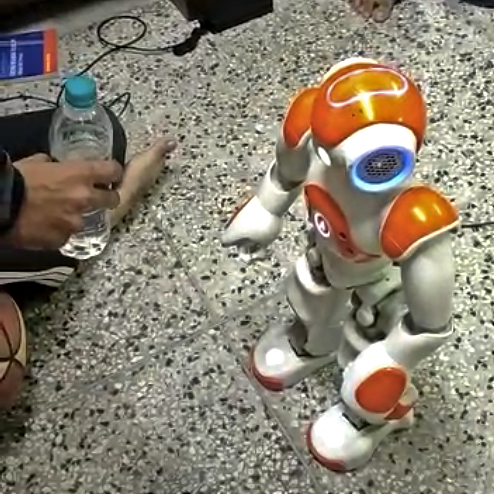}}
  \caption{(a), (b) Testing the inference model on the NAO robot to identify the primary object among distracting objects in view and take actions accordingly.}
  \label{exp}
\end{figure}

The neural network is fed the frame-by-frame image matrices to be processed it returns a list of $x$ and $y$ coordinates of the top-left and bottom-right corners of the bounding boxes along with the corresponding class label for each object detected. These labels are then be transferred back to the robot facilitating it to undertake programmable actions corresponding to the detected objects. In case the frame rate is too high, or the inference model throttles, five intermittent frames are dropped to keep up with the real-world environment and avoid returning older detected results to the robot.

\begin{figure*}[thpb]
    \centering
  \subfloat[\label{io-a}]{%
       \includegraphics[width=0.45\linewidth]{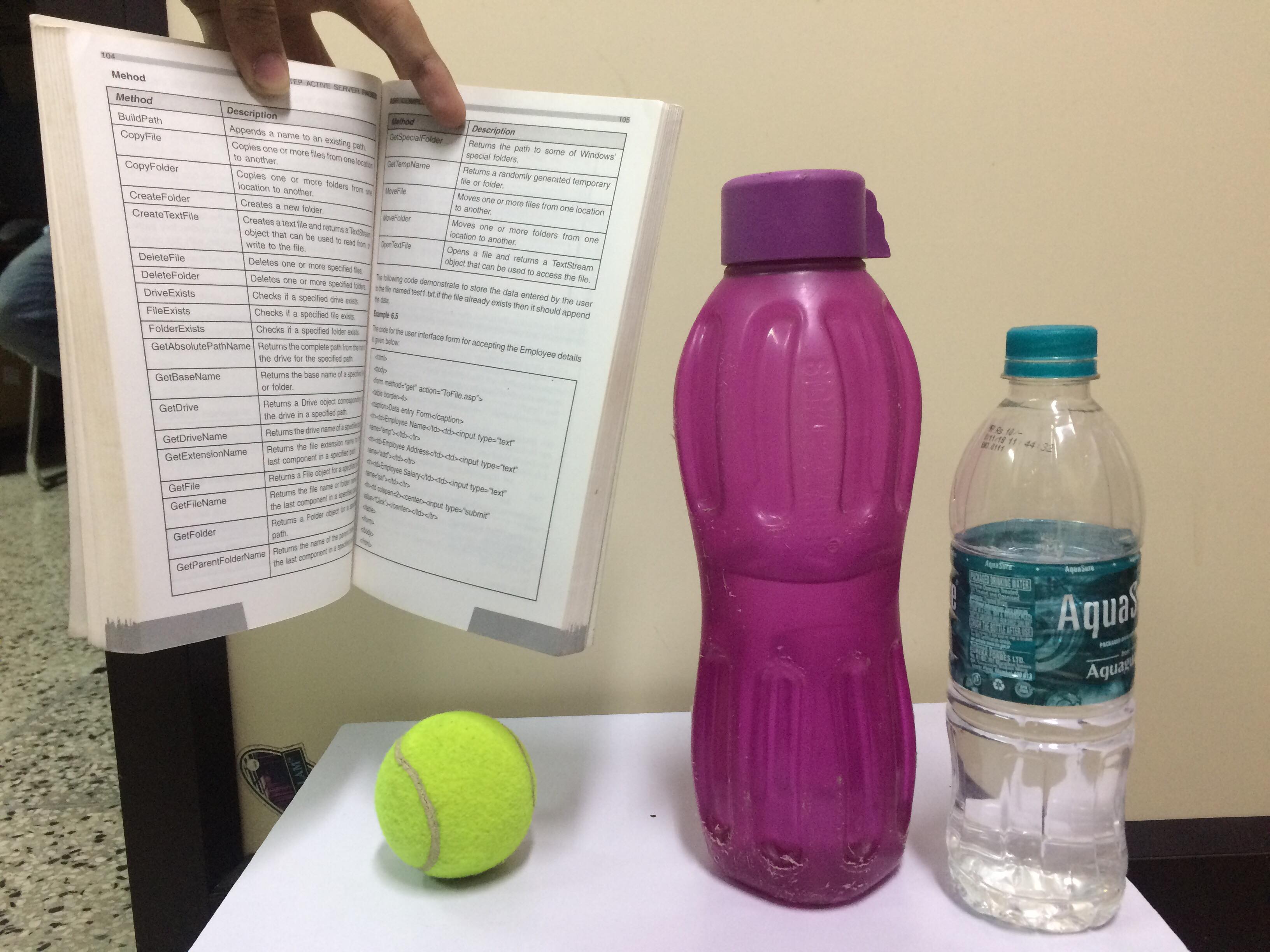}}
  \subfloat[\label{io-b}]{%
        \includegraphics[width=0.45\linewidth]{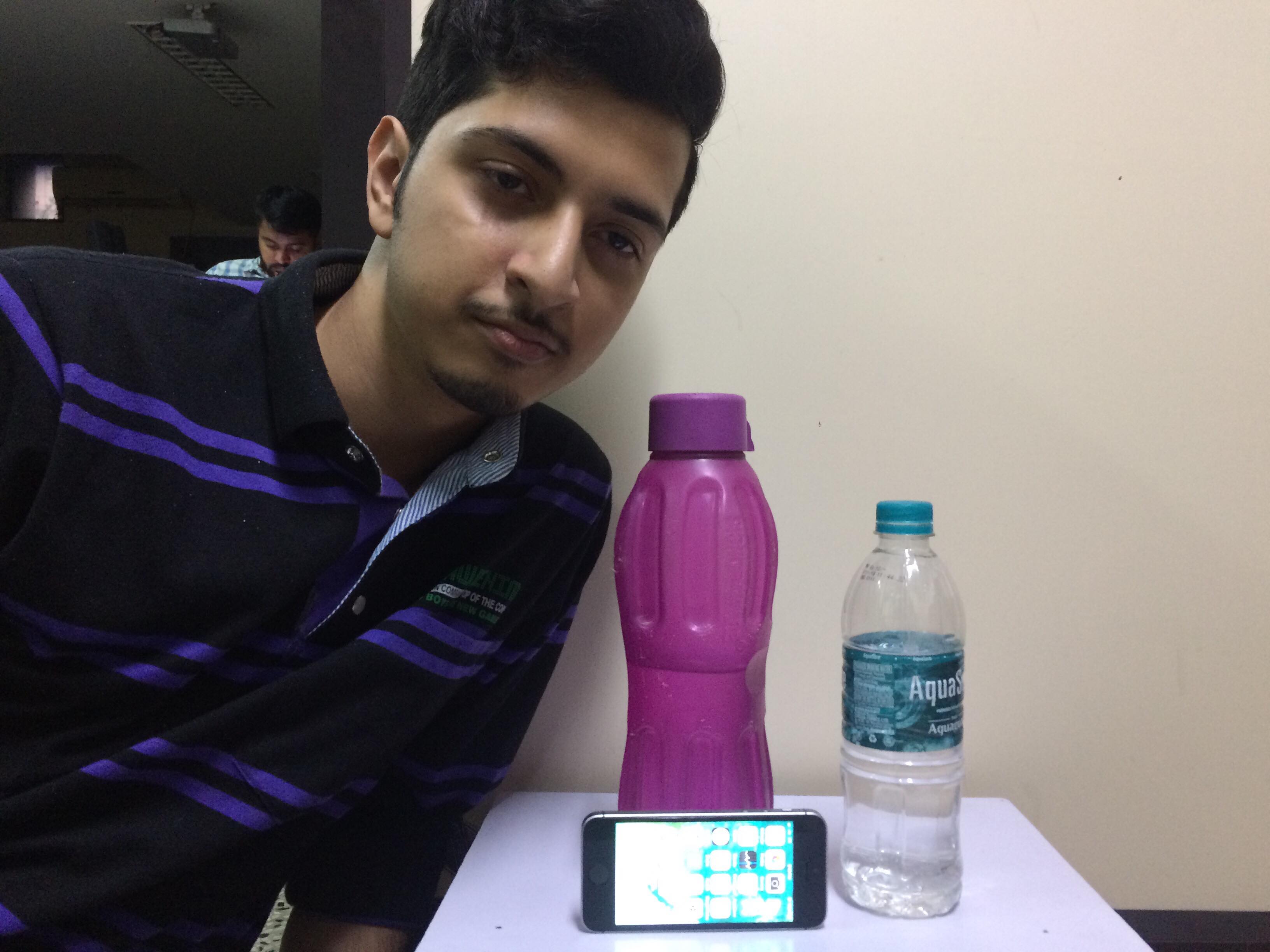}}
    \\
  \subfloat[\label{io-c}]{%
        \includegraphics[width=0.45\linewidth]{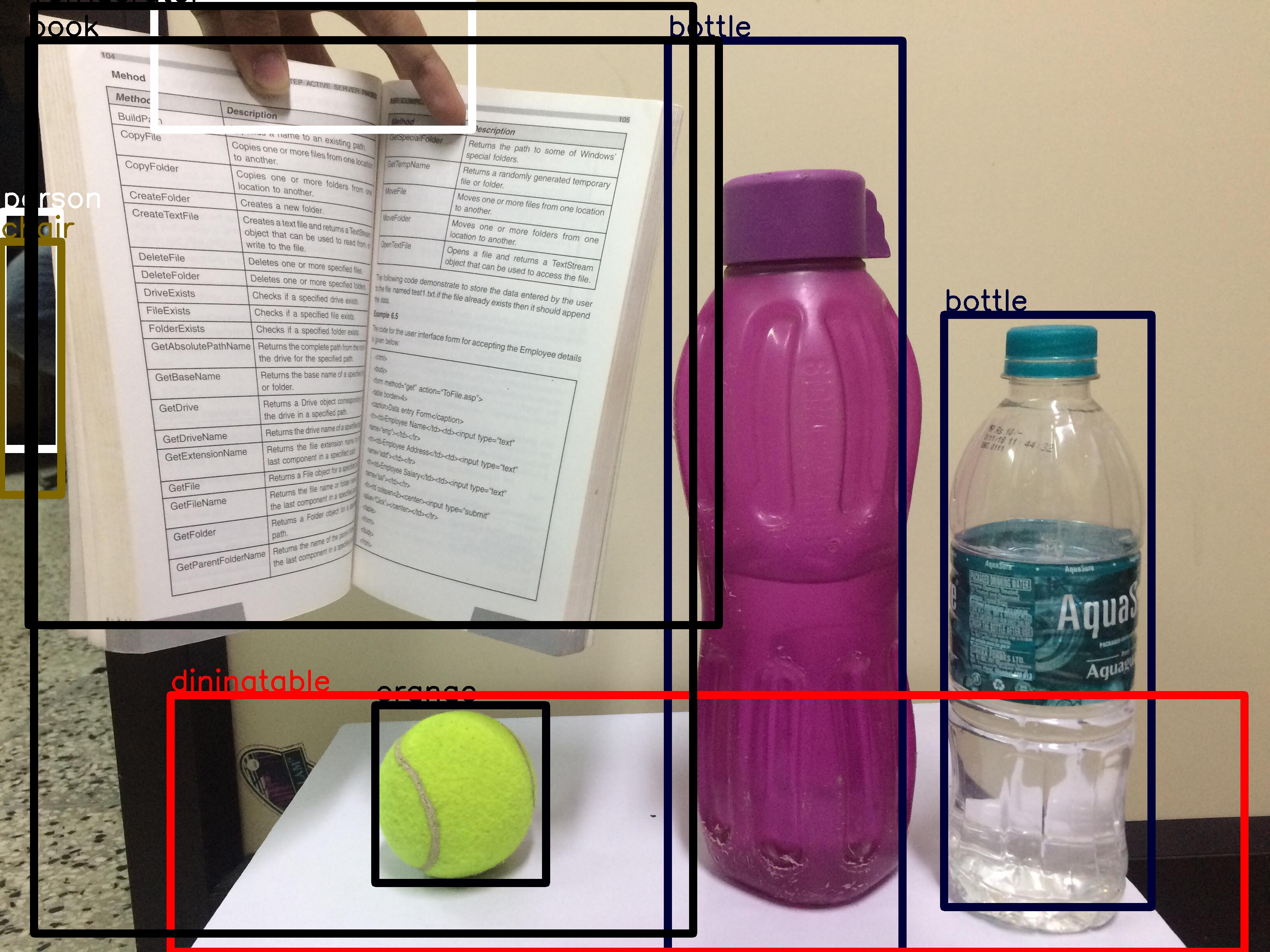}}
  \subfloat[\label{io-d}]{%
        \includegraphics[width=0.45\linewidth]{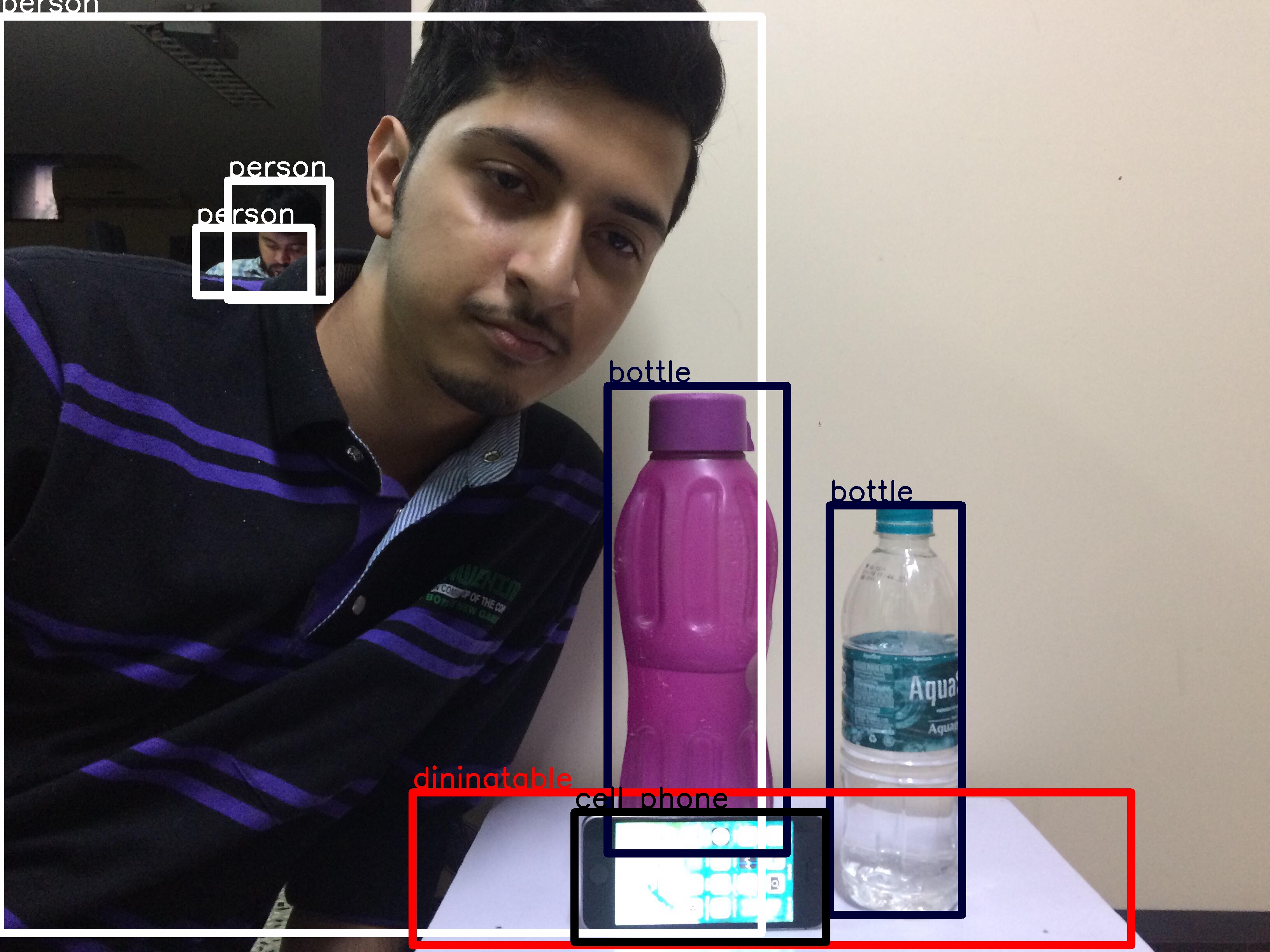}}
  \caption{(a), (b) Input frames from the NAO robot's camera feed.
  (c), (d) Corresponding output frames with bounding boxes and object labels.}
  \label{io}
\end{figure*}

Using the inference engine as a central control point, we have performed tests by connecting up to 3 robots operating in 3 different rooms in the laboratory environment and managed to communicate with all of them simultaneously without any additional latency. In case of multiple connections, the input frames are marked with the IP address of the source robot in the metadata and put in the queue and the bounding box results are sent back to the corresponding robot after processing. This approach, accompanied by the considerably low inference time, helps us in having multiple functional robots using a single deployment of the learning network, thereby making it computationally economic.
\section{Results and Performance Comparisons}
\label{results}

As per our experiments carried out in laboratory environment (Fig.~\ref{exp}), the proposed neural network model produces the results shown in Fig.~\ref{io} as observed from the monitoring stream and is able to successfully resolve multiple closely spaced objects in real-time.

\begin{figure}[htpb]
    \centering
    \includegraphics[width=\columnwidth]{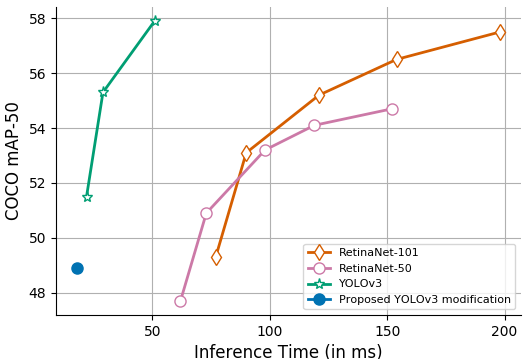}
    \caption{Results of YOLOv3 modification compared against state-of-the-art RetinaNet models and other YOLOv3 variants as a demonstration of the low inference time and at relatively at par accuracy of the proposed model}
    \label{comparison-graph}
\end{figure}

\begin{figure}[htpb]
    \centering
    \includegraphics[width=\columnwidth]{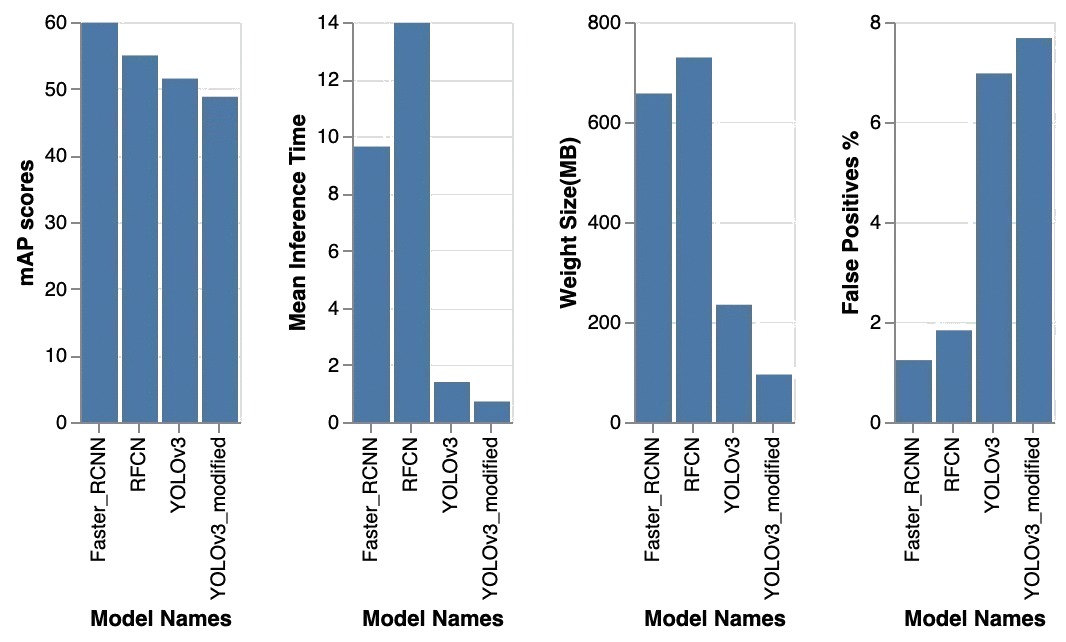}
    \caption{Comparative results of bounding box algorithms on COCO dataset alongside the modified model used in the NAO robot}
    \label{comparison}
\end{figure}

As demonstrated in Fig.~\ref{comparison}, using our modified YOLO architecture (weight size = 94.89MB), we have been able to achieve both sufficient accuracy ($mAP_{50}$ = 48.984 in comparison to 51.5 for YOLOv3-320, net loss = 4.88\%; mean inference time = 17 ms) in prediction on the COCO dataset as well as sub-second mean pipeline inference time of 0.7655s (from 1.328s for original YOLOv3, averaged over video streams of 120s) including the hardware latency for message transfer between the NAO robot and the inference engine. The detailed results in comparison to the existing state-of-the-art RetinaNet (high accuracy) and YOLOv3 (low latency) models have been shown in Table~\ref{comparison-table} and illustrated in Fig.~\ref{comparison-graph}. The available action list can be easily extended in future to add more human-like interactions.

\begin{center}
    \begin{table}[htpb]
        \caption{Detailed Performance Results on COCO}
        \label{comparison-table}
        \begin{tabular}{ccc}
            \toprule
            Method & $mAP_{50}$ & Inference Time (ms) \\
            \midrule
            RetinaNet-50-500 & 50.9 & 73 \\
            RetinaNet-101-500 & 53.1 & 90 \\
            RetinaNet-101-800 & 57.5 & 198 \\
            \midrule
            YOLOv3-320 & 51.5 & 22 \\
            YOLOv3-416 & 55.3 & 29 \\
            YOLOv3-608 & 57.9 & 51 \\
            \midrule
            Proposed YOLOv3 for NAO Robot & 48.9 & 17 \\
            \bottomrule
        \end{tabular}
    \end{table}
\end{center}

A separate test was carried out to check the performance of the modified neural network model in heavily crowded images, keeping in mind the primary limitation of YOLO, i.e., struggling to resolve individual small objects in a large group. In such circumstances, the modified model performs very similar (7.4\%) to the original YOLOv3 model (6.9\%) in terms of percentage of false positives, despite having lesser layers.
\section{Discussion And Future Work}
\label{discussion}

This paper studies the problem of enabling a low-compute humanoid robot like NAO to detect, recognize and localize objects in its field of view in real-time using the proposed modified YOLOv3 network model. Using an effective optimization to replace floating point calculations by integer manipulations and re-adjusting the convolutional layer specifications, the network becomes the integral part of an effective pipeline to receive the video frames from the NAO robot's camera and use the yielded results for the guidance of the latter. 

In terms of an environmental awareness of the robot, the main feature that can be incorporated in the object detection pipeline is exact localization in 3D space. This can be done with depth sensors as they have been useful in applications like real-time tele-walking of NAO \cite{nao_telewalking}. Our proposed solution is able to achieve fast and accurate detection results, which in turn leads to quicker action responses. However, the actions that are being taken currently are agnostic to the actual distance of the object from the robot. If a depth sensor is used then the robot can actually be moved to the exact location an object (like a sports ball) and the action can be performed directly on that object.

It is also necessary for the detection algorithm to perform better in resolving individual small objects from a dense cluster, which will open up the chances to add many more actions to the robot's available choices, e.g., picking up a marble from a bunch, etc.

The entire pipeline can be made even faster to improve the frames-per-second of the resultant output stream. Though this might not offer a significant improvement in the speed of actions being taken, it can provide a better output stream which looks more fluid.

As of the current implementation of a modified YOLO network, our proposed solution has been able to achieve a real-time detection and recognition of multiple objects in the robot's field of view, along with 2D localization. Incorporating the above mentioned feature enhancements and implementing the proposed system on a robot with better grasping capabilities will make the robot better fit to perform in a real-world environment.

Using the inference engine as a central hub has been proven to be a successful method of operating multiple robots within the network radius, e.g. a household environment, using a single deployment of the trained network. This engine may further be integrated with smart voice-controlled assistants to make the robots dynamically actionable on command with respect to the objects in its field of view, thereby forming a robust assistance system.

\addtolength{\textheight}{-12cm}   





\section*{Acknowledgments}

The authors gratefully acknowledge the Robotics and Machine Intelligence Laboratory at the Indian Institute of Information Technology, Allahabad for having granted access to perform the experiments with the NAO robots.



\bibliographystyle{IEEEtran}
\bibliography{bibliography}

\begin{thebibliography}{10}
\providecommand{\url}[1]{#1}
\csname url@rmstyle\endcsname
\providecommand{\newblock}{\relax}
\providecommand{\bibinfo}[2]{#2}
\providecommand\BIBentrySTDinterwordspacing{\spaceskip=0pt\relax}
\providecommand\BIBentryALTinterwordstretchfactor{4}
\providecommand\BIBentryALTinterwordspacing{\spaceskip=\fontdimen2\font plus
\BIBentryALTinterwordstretchfactor\fontdimen3\font minus
  \fontdimen4\font\relax}
\providecommand\BIBforeignlanguage[2]{{%
\expandafter\ifx\csname l@#1\endcsname\relax
\typeout{** WARNING: IEEEtran.bst: No hyphenation pattern has been}%
\typeout{** loaded for the language `#1'. Using the pattern for}%
\typeout{** the default language instead.}%
\else
\language=\csname l@#1\endcsname
\fi
#2}}

\bibitem{nao_autism}
S.~{Shamsuddin}, H.~{Yussof}, L.~{Ismail}, F.~A. {Hanapiah}, S.~{Mohamed},
  H.~A. {Piah}, and N.~I. {Zahari}, ``Initial response of autistic children in
  human-robot interaction therapy with humanoid robot nao,'' in \emph{2012 IEEE
  8th International Colloquium on Signal Processing and its Applications},
  March 2012, pp. 188--193.

\bibitem{nao_elderly}
\BIBentryALTinterwordspacing
D.~L\'{o}pez~Recio, E.~M\'{a}rquez~Segura, L.~M\'{a}rquez~Segura, and A.~Waern,
  ``The nao models for the elderly,'' in \emph{Proceedings of the 8th ACM/IEEE
  International Conference on Human-robot Interaction}, ser. HRI '13.\hskip 1em
  plus 0.5em minus 0.4em\relax Piscataway, NJ, USA: IEEE Press, 2013, pp.
  187--188. [Online]. Available:
  \url{http://dl.acm.org/citation.cfm?id=2447556.2447630}
\BIBentrySTDinterwordspacing

\bibitem{nao_elderly2}
J.~P.~M. {Vital}, M.~S. {Couceiro}, N.~M.~M. {Rodrigues}, C.~M. {Figueiredo},
  and N.~M.~F. {Ferreira}, ``Fostering the nao platform as an elderly care
  robot,'' in \emph{2013 IEEE 2nd International Conference on Serious Games and
  Applications for Health (SeGAH)}, May 2013, pp. 1--5.

\bibitem{nao_nonverbal_cues}
J.~{Han}, N.~{Campbell}, K.~{Jokinen}, and G.~{Wilcock}, ``Investigating the
  use of non-verbal cues in human-robot interaction with a nao robot,'' in
  \emph{2012 IEEE 3rd International Conference on Cognitive Infocommunications
  (CogInfoCom)}, Dec 2012, pp. 679--683.

\bibitem{detection}
\BIBentryALTinterwordspacing
R.~B. Girshick, J.~Donahue, T.~Darrell, and J.~Malik, ``Rich feature
  hierarchies for accurate object detection and semantic segmentation,''
  \emph{CoRR}, vol. abs/1311.2524, 2013. [Online]. Available:
  \url{http://arxiv.org/abs/1311.2524}
\BIBentrySTDinterwordspacing

\bibitem{coco}
T.-Y. Lin, M.~Maire, S.~Belongie, J.~Hays, P.~Perona, D.~Ramanan,
  P.~Doll{\'a}r, and C.~L. Zitnick, ``Microsoft coco: Common objects in
  context,'' in \emph{Computer Vision -- ECCV 2014}, D.~Fleet, T.~Pajdla,
  B.~Schiele, and T.~Tuytelaars, Eds.\hskip 1em plus 0.5em minus 0.4em\relax
  Cham: Springer International Publishing, 2014, pp. 740--755.

\bibitem{fast-rcnn}
\BIBentryALTinterwordspacing
R.~B. Girshick, ``Fast {R-CNN},'' \emph{CoRR}, vol. abs/1504.08083, 2015.
  [Online]. Available: \url{http://arxiv.org/abs/1504.08083}
\BIBentrySTDinterwordspacing

\bibitem{faster-rcnn}
\BIBentryALTinterwordspacing
S.~Ren, K.~He, R.~Girshick, and J.~Sun, ``Faster r-cnn: Towards real-time
  object detection with region proposal networks,'' in \emph{Proceedings of the
  28th International Conference on Neural Information Processing Systems -
  Volume 1}, ser. NIPS'15.\hskip 1em plus 0.5em minus 0.4em\relax Cambridge,
  MA, USA: MIT Press, 2015, pp. 91--99. [Online]. Available:
  \url{http://dl.acm.org/citation.cfm?id=2969239.2969250}
\BIBentrySTDinterwordspacing

\bibitem{mask-rcnn}
K.~He, G.~Gkioxari, P.~Doll{\'a}r, and R.~B. Girshick, ``Mask r-cnn,''
  \emph{2017 IEEE International Conference on Computer Vision (ICCV)}, pp.
  2980--2988, 2017.

\bibitem{rfcn}
\BIBentryALTinterwordspacing
J.~Dai, Y.~Li, K.~He, and J.~Sun, ``R-fcn: Object detection via region-based
  fully convolutional networks,'' in \emph{Proceedings of the 30th
  International Conference on Neural Information Processing Systems}, ser.
  NIPS'16.\hskip 1em plus 0.5em minus 0.4em\relax USA: Curran Associates Inc.,
  2016, pp. 379--387. [Online]. Available:
  \url{http://dl.acm.org/citation.cfm?id=3157096.3157139}
\BIBentrySTDinterwordspacing

\bibitem{yolo}
\BIBentryALTinterwordspacing
J.~Redmon, S.~K. Divvala, R.~B. Girshick, and A.~Farhadi, ``You only look once:
  Unified, real-time object detection,'' in \emph{2016 {IEEE} Conference on
  Computer Vision and Pattern Recognition, {CVPR} 2016, Las Vegas, NV, USA,
  June 27-30, 2016}, 2016, pp. 779--788. [Online]. Available:
  \url{https://doi.org/10.1109/CVPR.2016.91}
\BIBentrySTDinterwordspacing

\bibitem{yolov3}
\BIBentryALTinterwordspacing
J.~Redmon and A.~Farhadi, ``Yolov3: An incremental improvement,'' \emph{CoRR},
  vol. abs/1804.02767, 2018. [Online]. Available:
  \url{http://arxiv.org/abs/1804.02767}
\BIBentrySTDinterwordspacing

\bibitem{yolo9000}
J.~{Redmon} and A.~{Farhadi}, ``Yolo9000: Better, faster, stronger,'' pp.
  6517--6525, July 2017.

\bibitem{intent_object_grasping}
F.~H. Zunjani, S.~Sen, H.~Shekhar, A.~Powale, D.~Godnaik, and G.~C. Nandi,
  ``Intent-based object grasping by a robot using deep learning,'' in
  \emph{2018 IEEE 8th International Advance Computing Conference (IACC)}, In
  Press.

\bibitem{imagenet}
A.~Krizhevsky, I.~Sutskever, and G.~E.~Hinton, ``Imagenet classification with
  deep convolutional neural networks,'' \emph{Neural Information Processing
  Systems}, vol.~25, 01 2012.

\bibitem{guidelines}
N.~Cruz, K.~Lobos-Tsunekawa, and J.~R. del Solar, ``Using convolutional neural
  networks in robots with limited computational resources: Detecting nao robots
  while playing soccer,'' 2017.

\bibitem{yolo-quantized}
\BIBentryALTinterwordspacing
Y.~J. Wai, Z.~bin Mohd~Yussof, S.~I. bin Salim, and L.~K. Chuan, ``Fixed point
  implementation of tiny-yolo-v2 using opencl on fpga,'' \emph{International
  Journal of Advanced Computer Science and Applications}, vol.~9, no.~10, 2018.
  [Online]. Available: \url{http://dx.doi.org/10.14569/IJACSA.2018.091062}
\BIBentrySTDinterwordspacing

\bibitem{yolo-optimised}
J.~Ma, L.~Chen, and Z.~Gao, \emph{Hardware Implementation and Optimization of
  Tiny-YOLO Network}, 02 2018, pp. 224--234.

\bibitem{scalable}
D.~Erhan, C.~Szegedy, A.~Toshev, and D.~Anguelov, ``Scalable object detection
  using deep neural networks,'' 12 2013.

\bibitem{nao_telewalking}
I.~{Almetwally} and M.~{Mallem}, ``Real-time tele-operation and tele-walking of
  humanoid robot nao using kinect depth camera,'' in \emph{2013 10th IEEE
  International Conference on Networking, Sensing and Control (ICNSC)}, April
  2013, pp. 463--466.

\end{thebibliography}

\end{document}